\documentclass[letterpaper]{article} 
\usepackage{aaai24}  
\usepackage{times}  
\usepackage{helvet}  
\usepackage{courier}  
\usepackage[hyphens]{url}  
\usepackage{graphicx} 
\usepackage{natbib}  
\usepackage{caption} 
\usepackage{subfigure}
\usepackage[table]{xcolor}
\usepackage{multirow}
\usepackage{booktabs}
\usepackage[ruled,linesnumbered]{algorithm2e}
\usepackage{algorithmic}
\usepackage{newfloat}
\usepackage{listings}
\DeclareCaptionStyle{ruled}{labelfont=normalfont,labelsep=colon,strut=off} 
\title{Recap: Detecting Deepfake Video with Unpredictable Tampered Traces via  Recovering Faces and Mapping Recovered Faces}
\author{
Juan Hu \textsuperscript{\rm 1},
Xin Liao \textsuperscript{\rm 1},
    Difei Gao \textsuperscript{\rm 2},
    Satoshi Tsutsui \textsuperscript{\rm 3},
    Qian Wang \textsuperscript{\rm 4},
    Zheng Qin \textsuperscript{\rm 1},
    Mike Zheng Shou\textsuperscript{\rm 2}\\
}
\affiliations{
    \textsuperscript{\rm 1} College of Computer Science and Electronic Engineering, Hunan University, China\\
\textsuperscript{\rm 2} Show Lab, National University of Singapore, Singapore\\
\textsuperscript{\rm 3} ROSE Lab, Nanyang Technological University, Singapore\\
\textsuperscript{\rm 4} School of Cyber Science and Engineering, Wuhan University, China\\
}

\usepackage{bibentry}

\begin{document}

\maketitle

\begin{abstract}

The exploitation of Deepfake techniques for malicious  intentions has driven significant research interest in Deepfake detection.
Deepfake manipulations frequently introduce random tampered traces,  leading to  unpredictable outcomes in different facial regions. However, existing detection methods heavily rely on specific forgery indicators,  and as the forgery mode improves, these traces become increasingly randomized, resulting in a decline in the detection performance of methods reliant on specific forgery traces. 
To address the limitation,  we propose \textsf{Recap}, a novel Deepfake detection model that exposes unspecific facial part inconsistencies by recovering faces and enlarges the differences between real and fake by mapping recovered faces. 
In the recovering stage, the model focuses on  randomly masking  regions of interest (ROIs)  and  reconstructing  real faces without unpredictable tampered traces, resulting in a relatively good recovery effect for real faces while a poor recovery effect for fake faces. In the mapping stage,  the output of the recovery phase serves as supervision to guide the facial mapping process. This mapping process  strategically emphasizes the mapping of fake faces with poor recovery, leading to a further deterioration in their representation, while enhancing and refining the mapping of real faces with good representation. As a result,  this approach significantly amplifies the  discrepancies between real and fake videos. Our extensive experiments on standard benchmarks demonstrate that \textsf{Recap} is effective in  multiple scenarios. 

\end{abstract}

\section{Introduction}\label{sec:intro}
Deepfakes, AI-generated videos of individuals like fabricated speech, pose serious threats to society~\cite{chesney2019deep,aibase}, emphasizing the need for reliable detection methods. While some detection methods focus on specific forgery patterns, state-of-the-art Deepfake techniques generate different forgeries in varying regions of a facial image, as depicted in Fig. \ref{fig1a} \cite{ffdata,wild,dfdc,celeb}. This variability results in unpredictable tampered traces, as shown in Fig. \ref{fig1b}, making it difficult to anticipate the traces in the next frame based on previous ones or from one facial area to another.  This unpredictability complicates Deepfake detection, as methods relying on specific patterns can be easily evaded. In real-world scenarios, videos may be generated using unknown manipulation techniques, rendering tampered traces even more unpredictable. Thus, it is crucial to develop methods that can detect unpredictable tampered traces in unknown domains.

\begin{figure}[!t]
 \centering
\includegraphics[width=\linewidth]{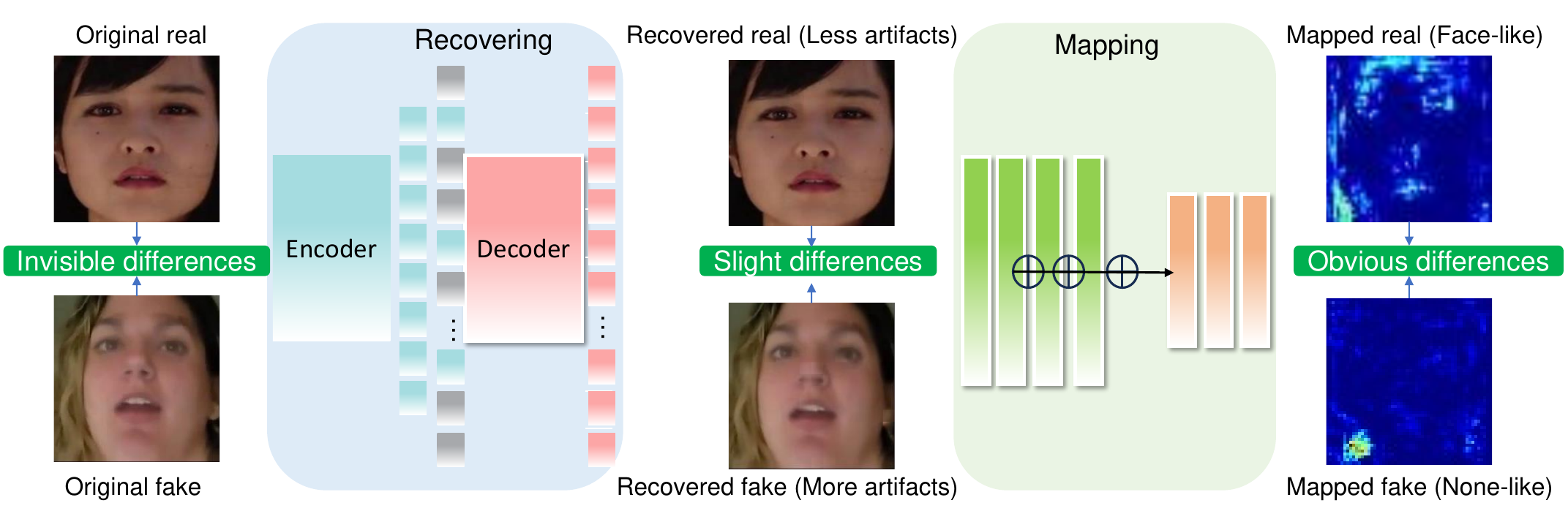}
   \vspace{-0.5cm}
    \caption{Recovering results and mapping results. The fake faces exhibit worse results than that of real faces after recovering. The mapping process maps poorly recovered fake faces into even worse ones, which magnifies the differences between real faces and fake faces.}
 \label{fig1}
\vspace{-0.5cm}
\end{figure}

Recently, reconstruction-prediction-based methods have achieved relatively high detection performance. These methods typically involve the encoder and  decoder -- encoding the  input data into a low-dimensional representation and subsequently decoding the original inputs from that representation. For example, \cite{khalid2020oc} use reconstruction scores to classify real and fake videos. Moreover, reconstructing and predicting future frame representations \cite{finfer}, forgery configurations \cite{self}, pseudo training samples \cite{chen2022ost},  artifact representations \cite{dong2022explaining},  the whole faces \cite{recon}, the masked relation \cite{yang2023masked}, and mask regions \cite{chen2022learning} can boost the detection performance. However, these methods overlook the importance of reprocessing the reconstructed results to amplify differences, which hinders the effective extraction of Deepfake clues.



\noindent\textbf{Our Ideas}. We present two relatively under-explored ideas for reconstruction-prediction-based Deepfake detection methods. First, according to \cite{wang2023noise}, tampered traces are left within the face area regardless of different forgery modes. To exploit the unpredictable nature of tampered traces in Deepfake videos, we design a unique masking strategy for a ``Recovering'' (Fig. \ref{fig1} left) module, amplifying the artifacts in Deepfake faces more as illustrated in the middle images of Fig. \ref{fig1}. Second, to further increase the difference between real and fake videos effectively, we design a novel ``Mapping'' (Fig. \ref{fig1} right) module that maps recovered real faces into a better representation but recovered fake faces into a worse representation, as illustrated in the right images of Fig. \ref{fig1}.

We name our method as \textsf{Recap}, which, in essence, works as follows: It begins with a real face image and pretrains a masked autoencoder. This autoencoder is guided by the facial parts, as outlined in Algorithm~\ref{algorithm1}, which allows it to highlight the inconsistencies in the unspecified facial parts of fake faces. Subsequently, the masked autoencoder predicts the masked regions of interest (ROIs) based on the unmasked facial parts and interframes. This strengthens the understanding of relationships between facial parts and their temporal consistency. To further enhance the detection performance, we introduce a model designed to map the recovered faces. As seen in Fig. \ref{fig1}, the goal of this mapping process is to ensure successful mapping for reconstructed real faces, while causing the failure to map fake ones.


\noindent\textbf{Contributions}. 
\noindent{(1)} We propose  \textsf{Recap} to learn unspecific features among all facial parts, enabling the detection of Deepfake videos with unpredictable tampered traces. This method provides a crucial clue in identifying previously unknown Deepfakes. 

\noindent{(2)} To further maximize the discrepancy between real and fake videos,  we utilize the reconstructed results for mapping. This process ensures that well-reconstructed real faces are further mapped well, while poorly-reconstructed fake faces are mapped less effectively. 

\noindent{(3)}  Extensive experiments on benchmark datasets, including FaceForensics++ \cite{ffdata}, Celeb-DF \cite{celeb}, WildDeepfake (WildDF) \cite{wild}, and DFDC \cite{dfdc} show that \textsf{Recap} achieves effective performance under various metrics.

\begin{figure*}[t]
  \centering
   \includegraphics[width=0.9\linewidth]{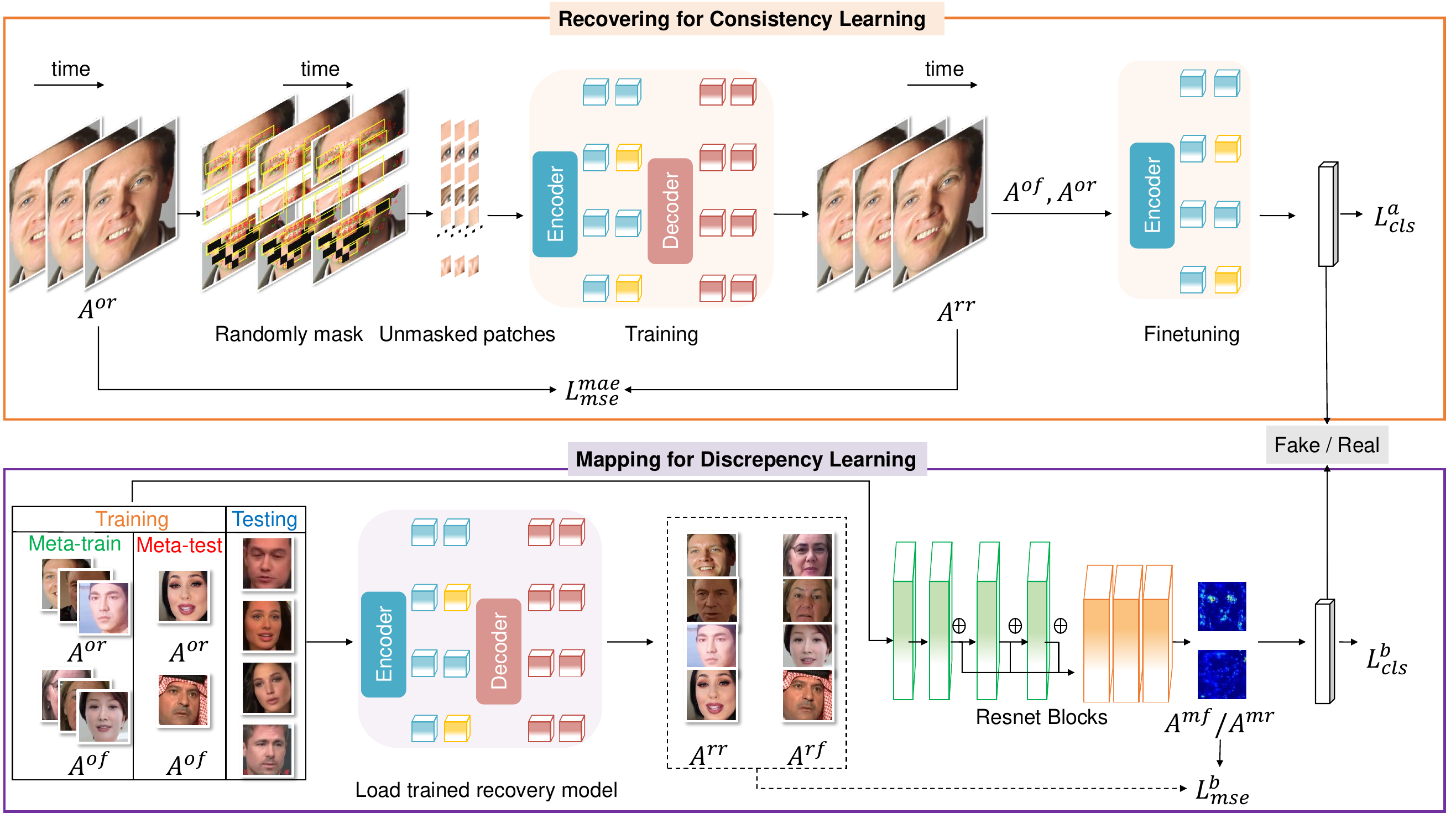}
\vspace{-0.3cm}
   \caption{Pipeline of the propose \textsf{Recap}. In the Recovering stage, \textsf{Recap} learns unspecific features by developing the designed masking strategy and recovery process. In the Mapping stage, \textsf{Recap}  maximizes the discrepancy between real videos and Deepfake videos by leveraging the recovered results to generate face-like maps of real faces and none-like maps of fake ones.}
   \label{fig2}
\vspace{-0.5cm}
\end{figure*}

\section{Related Work}
Before the emergence of the reconstruction-prediction-based methods explained in Sec.~\ref{sec:intro}, most detection approaches employed generalized methods. These methods can be broadly classified into three categories: those based on implicit clues, those based on explicit clues, and those that combine implicit and explicit clues. Methods based on implicit clues~\cite{meso, capsule,tan2019efficientnet, ffdata, smil, ltw, CNN-GRU, multiatt, aaaidual, aaailocal} use supervised learning to distinguish genuine and fake videos without explicitly incorporating clues to detect Deepfake videos, making it challenging to understand the underlying detection clues. Methods that employ explicit clues~\cite{eye, head, thinking, frequency, cnn2, emotions, luo2021generalizing,on, hcil, xray, lips, spsl, finfer, stil, UIA-ViT, Patch-based, pcl, ftcn, delv, haliassos2022leveraging,dong2022protecting,shiohara2022detecting, wang2023noise} have achieved more promising performance. Furthermore, Huang et al. explore explicit and implicit embeddings for Deepfake detection. However, given the rapid advancement of Deepfake technology, various falsification traces can be left behind, rendering detection methods that rely on specific facial features vulnerable to attack. 

\section{Method}
\label{sec:Method}
This section presents the details of \textsf{Recap} for Deepfake video detection.  Specifically, the proposed method is composed of two stages: (1) Recovering for the Consistency Learning, and (2) Mapping for the Discrepancy Learning stage, as shown in Fig.~\ref{fig2}. 

\noindent\textbf{Notations. }
Let $A^{or}$, $A^{of}$, $A^{rr}$, $A^{rf}$, $A^{mr}$, and $A^{mf}$ be original real faces, original fake faces, recovered real faces, recovered fake faces,  masked real faces, and masked fake faces.

\subsection{Recovering for Consistency Learning}\label{sec:mafpc}

In this stage, we perform self-supervised learning of real faces to learn generic facial part consistency features. As a result,  the unspecific inconsistencies of fake faces with unpredictable tampered traces are exposed. Furthermore, we finetune the model with real faces and fake faces.

\noindent \textbf{Masking strategy tailored to learn the consistent face representation. }We design a facial part masking strategy to ensure that the model can learn the consistencies of all facial parts. We show the pseudo-code in Algorithm \ref{algorithm1}, which takes the inputs of videos' faces and outputs the masked faces. Following the temporal masking of VideoMAE \cite{videomae} and modifying the frame level masking strategy of VideoMAE \cite{videomae}, we enforce different frames share the same masking regions to ensure the mask expands over the entire temporal axis. 

\indent The designed facial part masking strategy is different from the frame masking strategy of VideoMAE \cite{videomae}. First,  since the tampered traces may only be sporadically present in one part and not related to other facial parts, we devise the masking strategy by considering  Deepfake's domain knowledge. Specifically, we split the faces into different facial parts, i.e., eyes, cheek \& nose, and lips, enabling the model to focus on both local and global consistencies among all facial parts. We make this design choice because randomly masking pixels might disregard the importance of global consistency across various facial parts. Neglecting such global facial part consistency could impede the model's ability to learn accurate facial part consistency features, leading to difficulties in detecting differences between real and fake videos from reconstructed faces. 
{\renewcommand\baselinestretch{1.0}\selectfont

\begin{algorithm}[!t]

\caption{Masking process of facial parts}
\label{algorithm1}

\KwIn{Original faces: $A^{of}$, $A^{or}$,  Mask ratio: $M_r$;}

\KwOut{Masked  faces: $A^{mf}$, $A^{mr}$ ;}

Resize the face image with a size of 224*224, and detect $68$ keypoint landmarks in faces;

Define $ROIs=\{R_1,R_2...R_{11}\}$.

Define $196$ blocks of face image;

$M =\{M_1, M_2, M_3\} $, $M_1:$ eyes patches $\in$ \{$(0,0)$  $\rightarrow$  $17^{th}_{landmark}$\}, and $M_2:$ nose \& cheek patches $\in$ \{end of eyes patches   $\rightarrow$  $16^{th}_{landmark}$\}, and $M_3:$ lips patches $\in$ \{end of nose \& cheek patches   $\rightarrow$ the last patches of $A^{of}$, $A^{or}$\};

 \For {$i = 1$ \textbf{to} $3$}
    {Randomly select $M_i$;
    
     \For {$block \in selected$ \indent $M_{i}$}
     {\If {$ROI$ \indent$landmarks \in block$}
    {Sign the blocks $S_{blk}$; }
    { Number of mask patches  = $S_{blk}$ $\times$ $M_r$;
     
    Random mask the patches; }}}

\end{algorithm}

\par}
\indent Second, the original masking strategy of VideoMAE \cite{videomae}, with a high masking ratio, would make it too challenging to restore the original appearance without any artifacts or distortions. If reconstruction artifacts occur, real faces will contain them, and fake faces will display both reconstructed artifacts and tampering artifacts. This makes it difficult to distinguish between real and fake videos since both have artifacts. Therefore, we propose a masking strategy that focuses on ROIs and utilizes a relatively low masking ratio to enable the model to reconstruct the original faces more accurately. The ROIs extraction is partially inspired by Facial Action Coding System (FACS) \cite{facs}, which considers the action units of FACS as fundamental elements. Drawing from psychology studies \cite{facs, mico1, micro2, expression, xinli}, it is well-known that real faces exhibit inherent consistency in these elements. Consequently, when we mask ROIs, it becomes more challenging to reconstruct these regions for fake faces compared to real faces.
We reference the action units of eyebrows,  lower eyelid, nose root, cheeks, mouth corner, side of the chin, and  chin to calculate bounding box coordinates according to facial keypoints. Specifically,  $R_1$ and $R_2$ are the eyebrows, and $R_3$ and $R_4$ are the lower eyelid, and $R_5$ is the nose root, and $R_6$ and $R_7$ are the cheeks, and $R_8$ and $R_9$ are the mouth corner, and $R_{10}$ is the side of the chin, and $R_{11}$ is the chin.  We discuss more about the masking strategy in the ablation study. 

\vspace{-0.05cm}
 \noindent \textbf{Network architecture.}  Our masked autoencoder is based on an asymmetric encoder-decoder architecture \cite{imgmae}. To consider temporal correlation, the vanilla  Vision Transformers (ViT)  and joint space-time attention \cite{videomae} are adopted for recovering.
 \\
   \textbf{Recover masked faces.} The masked patches of faces are dropped in the processing of the encoder, leaving the unmasked areas. In this way, the decoder predicts the missing facial part based on the unmasked areas. The reconstruction quality of masked patches is calculated with the MSE loss function $L^{mae}_{mse}$. If the model learns consistencies among facial parts, the loss between the reconstructed patches and the input patches should decrease. Our facial part masking strategy makes each part selected randomly, which enforces the model to learn the representation unspecific to any facial part. Furthermore, because this stage only uses real videos and does not use any Deepfake videos, it can prevent the model from over-fitting to any specific tampering pattern. In this way, the pretrained recovery model is obtained. Let  $A^{rr} = RE^r \times A^{or}, \mbox{subject  to } 0<RE^r<1,$ where $RE^r$ represents the recovery quality of $A^{or}$.  A higher score of $RE^r$ indicates better reconstruction quality.
\begin{figure}[!t]

\centering
\subfigure[Different forgery patterns]
{
\includegraphics[width=3.9cm]{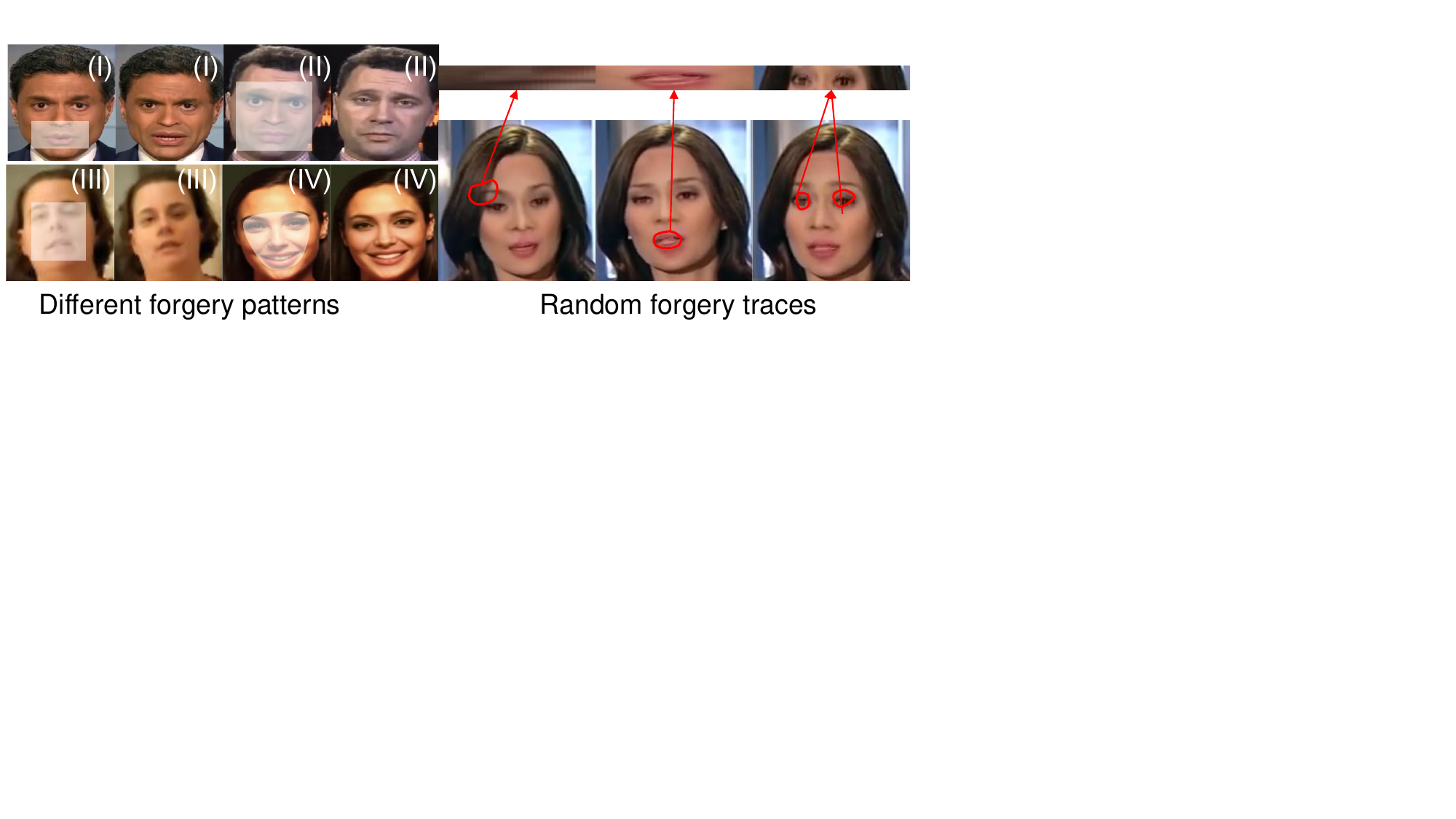}
\label{fig1a}
}
\subfigure[Random forgery traces]
{
\includegraphics[width=3.9cm]{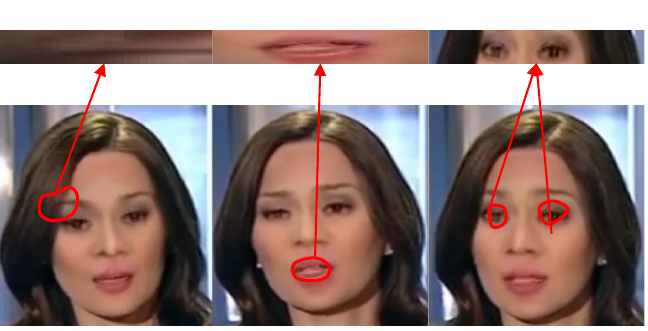}
\label{fig1b}
}
\vspace{-0.4cm}
\caption{Different forgery patterns use different shapes to replace the face area, making the tampered traces random and unpredictable. (I) Face2Face in FaceForensics++ \cite{ffdata}. (II) FSGAN in WildDeepfake (III) Deepfake in DFDC \cite{dfdc}. (IV) Deepfake in Celeb-DF \cite{celeb}.}
\label{figab}
\vspace{-.2in}
\end{figure}

\noindent \textbf{Finetuning the recovery model. }We discard the decoder and apply the encoder to uncorrupted $A^{or}$  and  $A^{of}$  for finetuning. The finetuning process uses a cross-entropy loss $L_{cls}^{a}$ for detection. Since the recovery model learns the facial part consistency of real videos, the well-trained encoder can extract the consistency features of real videos. For fake videos, as shown Fig. \ref{figab}, they are generated by different forgery patterns and tampered with different areas, and the tampered traces can show up in random regions. Consequently, the tampered traces can not be predicted. If the masked areas contain tampered traces, the recovery process can be affected. If there are no tampered traces in the masked area, the  tampered traces in unmasked areas can not be recovered. That is, regardless of whether the tampering traces are covered, the video with unpredictable tampering traces will influence the recovery process, which makes the features extracted from the encoder  different from that of the original videos.

\subsection{Mapping for Discrepancy Learning}\label{sec:stage2}
In this stage, we leverage the well-trained recovery model from the first stage and map the recovery result to enlarge the discrepancy between real and fake videos.

\noindent\textbf{Input data. }We load the trained recovery model to obtain $A^{or}$, and input  $A^{of}$, $A^{or}$, $A^{rf}$, and $A^{rr}$ into the Mapping stage. Let $A^{rf} = RE^f \times A^{of}$, where $RE^f$ represents the recovery quality of $A^{of}$.

\noindent \textbf{Data split strategy.} To avoid over-fitting to specific Deepfake patterns, we use meta-learning \cite{meta} and randomly divide the training data into Meta-train set and Meta-test set, where fake faces in Meta-train and Meta-test have different manipulated patterns.  \\
   \textbf{Network architecture. } We utilize the first convolutional layer of ResNet-18 \cite{resnet}.  The first three residual blocks of ResNet-18 are employed, and the outputs of these residual blocks are concatenated. The concatenated outputs are fed into three convolutional layers for face mapping. The dimensions of the mapped faces are $56 \times 56 \times 3$.  In this way, $A^{mr}$ and $A^{mf}$ can be represented as, $A^{mr} = MA^r \times A^{rr}, \mbox{subject  to } 0<MA^r<1$, $A^{mf} = MA^r \times A^{rf}, \mbox{subject  to } 0<MA^f<1$, where $MA^r$ and $MA^f$ represent the mapping quality of $A^{rr}$ and $A^{rf}$. 
 Ultimately, a fully connected layer is utilized for classifications.  
\noindent\textbf{Detection loss. } To amplify the differences between $A^{or}$ and $A^{of}$, we should satisfy:
\begin{equation}
A^{mr}-A^{mf} \gg A^{or}-A^{of}.
\label{mloss}
\end{equation}
Combine the analyses of the $A^{mr}$, $A^{mf}$, $A^{or}$, and $A^{of}$, Eq. (\ref{mloss}) can be represented as:
\begin{equation}
A^{rr}(MA^r - \frac{1}{RE^{r}}) \gg A^{rf}(MA^f - \frac{1}{RE^{f}}).
\label{cb1}
\end{equation}
Since the  recovery model is trained on real data $A^{or}$ and the unpredictable traces of $A^{of}$ could influence the recovery process, we have $A^{rr}>A^{rf} $. Moreover, the recovery quality of $A^{of}$ can be smaller than that of $A^{or}$. That is, $0<RE^f<RE^r<1$. To satisfy Eq. (\ref{cb1}),  it is necessary to ensure that $MA^r \gg MA^f $. Therefore, we minimize the MSE loss $L_{mse}^{b}$ between the mapped faces and the reconstructed faces. Consequently, $L_{mse}^{b}$ allows the $A^{mr}$ to be constrained by the consistency of $A^{rr}$, while the $A^{rf}$ are constrained by the inconsistency. Ultimately, the model is able to recover $A^{rr}$ but fails to recover $A^{rf}$, ensuring that $MA^r \gg MA^f $.  Moreover, we also minimize the binary cross-entropy $L_{cls}^{b}$ between the video label and the output of the fully connected. 

For each epoch, a sample batch is formed with the same number of fake videos and real videos to construct the binary detection task. To simulate unknown domain detection during training, the Meta-train phase performs training by sampling many detection tasks, and is validated by sampling many similar detection tasks from the Meta-test. Thereafter, the parameters of Meta-train phase can be updated. The goal of Meta-test phase is to enforce a classifier that performs well on Meta-train and can quickly generalize to the unseen domains of Meta-test, so as to improve the cross-domain detection performance.

The final loss function of the Mapping Stage is:
\begin{equation}
L^b =  (  L^b_{cls} + L^b_{mse} )_{Meta-train}+( L^b_{cls} + L^b_{mse} )_{Meta-test}.
\end{equation}
which combines the Meta-test loss of $L_{cls}^{b}$ and $L_{mse}^{b}$ and Meta-train loss of $L_{cls}^{b}$ and $L_{mse}^{b}$ to achieve joint optimization.

\noindent\textbf{Detection results. }We average the output of Recovering  stage and Mapping stage to get the final detection score.

\begin{table*}[!htbp]
\begin{center}
\vspace{-0.4cm}
\caption{Comparisons of detection performance  (AUC (\%) and EER (\%)) between \textsf{Recap} and other methods on Celeb-DF, WildDF, and DFDC datasets when trained on $4$ types of videos of FaceForensics++. }
\vspace{-0.3cm}
\label{crossdatasets}
\renewcommand{\tabcolsep}{4mm} 
\renewcommand{\arraystretch}{1.2}
\begin{tabular}{ccccccccc}
\noalign{\hrule height 0.7pt}


     \multirow{2}{*}{Method}
     &\multicolumn{2}{c} {Celeb-DF}&\multicolumn{2}{c} {WildDF}& \multicolumn{2}{c} {DFDC}&\multicolumn{2}{c} {Avg}\\ 
    \cmidrule(r){2-3} \cmidrule(r){4-5}\cmidrule(r){6-7}\cmidrule(r){8-9}&AUC $\uparrow$ &EER$\downarrow$ &AUC $\uparrow$ &EER$\downarrow$&AUC $\uparrow$ &EER$\downarrow$&AUC $\uparrow$ &EER$\downarrow$\\
     
              \noalign{\hrule height 0.7pt}


{\textmd{MultiAtt}}&$76.7$&$32.8$&$70.2$&$36.5$& $67.3$&$38.3$&$71.4$&$35.9$ \\
{\textmd{LipForensics}}&$82.4$&$24.2$&$78.9$&$29.1$& $73.5$&$36.5$&$78.3$&$29.9$ \\
{\textmd{Huang et al.}}&$83.8$&$24.9$&$80.8$&$23.3$& $81.2$&$26.8$&$81.9$&$25.0$ \\
\hline
{\textmd{FInfer}}&$70.6$&$34.8$&$76.2$&$32.1$& $70.9$&$36.4$&$72.6$&$34.4$ \\
{\textmd{SSL-AE}}&$79.7$&$30.8$&$76.1$&$32.9$& $72.3$&$35.8$&$76.0$&$33.2$ \\
{\textmd{OST }}&$74.8$&$31.2$&$80.1$&$28.7$& $\textbf{83.3}$&$25.0$&$79.4$&$28.3$ \\

{\textmd{RECCE}}&$73.7$&$30.3$&$70.3$&$35.5$& $74.0$&$31.1$&$72.7$&$32.3$ \\
{\textmd{MRL}}&$86.7$&$18.3$&$70.1$&$35.6$& $74.5$&$30.1$&$77.1$&$28.0$ \\
 \rowcolor{gray!20}
{\textsf{Recap}}
&${\textbf{90.1}}$&$\textbf{{14.2}}$&${\textbf{83.1}}$&${\textbf{24.6}}$&${{83.2}}$&$\textbf{24.8}$&${\textbf{85.5}}$&$\textbf{21.2}$\\

\noalign{\hrule height 0.7pt}

\end{tabular}
\end{center}
\vspace{-0.5cm}
\end{table*}

\section{Experiment}
\label{sec:Experiment}
\subsection{Experimental Setup}
\indent \textbf{Datasets.} Four public Deepfake videos datasets, i.e.,  FaceForensics++ \cite{ffdata}, Celeb-DF  \cite{celeb}, WildDF \cite{wild}, DFDC \cite{dfdc} are utilized to evaluate the proposed method and existing methods. FaceForensics++ is made up of $4$ types manipulated algorithms: DeepFakes \cite{deepfakegit},  Face2Face \cite{thies2016face2face}, FaceSwap \cite{faceswapgit}, NeuralTextures \cite{ne}. Moreover, $4000$ videos are synthesized based on the $4$ algorithms. These videos are widely used in various Deepfake detection scenarios. Celeb-DF contains $5639$ videos that are generated by an improved DeepFakes algorithm \cite{celeb}. The tampered traces in some inchoate datasets are relieved in Celeb-DF. WildDF consists of $707$ Deepfake videos that were collected from the real world. The real world videos contain diverse scenes, facial expressions, and forgery types, which makes the datasets challenging. DFDC   is a large-scale Deepfake detection dataset published by Facebook.

\noindent \textbf{Implementation details.} In the Recovering stage, the masking ratio, batch size, patch size, and input size are set as $0.75$, $8$, $16$, $224$, respectively. The AdamW \cite{adamw} optimizer with an initial learning rate $1.5 \cdot 10^{-4}$, momentum
of $0.9$ and a weight decay $0.05$ is utilized to train the recovery model. The finetuning of the Recovering stage utilizes the AdamW  optimizer with an initial learning rate $1 \cdot 10^{-3}$ to detect videos. The
SGD optimizer is used for optimizing the Mapping stage with the initial learning rate $0.1$, momentum of $0.9$, and weight decay of $5\cdot 10^{-4}$. We use \texttt{FFmpege} \cite{ffmpeg} to extract $30$ frames from each video. The \texttt{dlib} \cite{dlib} is utilized to extract  faces and detect 68 facial landmarks. We randomly mask facial parts according to Algorithm \ref{algorithm1}. 

\noindent\textbf{Comparison methods. }
We
compare \textsf{Recap} with the reconstruction-prediction-based methods, i.e.,  FInfer \cite{finfer}, SSL-AE \cite{self}, OST \cite{chen2022ost},   RECCE \cite{recon}, and MRL \cite{yang2023masked}. We also compare \textsf{Recap} with generalized methods that are representative of   implicit methods, explicit methods, and explicit and implicit combined methods, i.e., MultiAtt \cite{multiatt}, LipForensics \cite{lips}, Huang et al. \cite{huang2023implicit}.

\subsection{Generalization to Unknown Domains}
 We enforce \textsf{Recap} to learn unspecific features for Deepfake video detection with unpredictable tampered traces. The unknown domain detection is precisely the scenario where tampered traces are often unpredictable. To test the performance of  \textsf{Recap}, we simulate unknown domain Deepfake detection in multiple scenarios.

First, we conduct experiments by training the model on FaceForensics++ with all $4$ types of videos, but testing on other datasets, i.e., Celeb-DF, WildDF, DFDC, and we use Area Under Curve (AUC) and Equal Error Rate (EER) to evaluate the performance.  The enormous differences between the training domain and the testing domain make it challenging to improve unknown domain detection performance. Nonetheless, the results in Table \ref{crossdatasets} show that  \textsf{Recap} manages to improve the performance by an average of 3.6\% AUC and 3.8\% EER.




\begin{table*}[!t]
\begin{center}
\vspace{-0.8cm}
\caption{Comparisons of the detection performance  (AUC (\%)) between \textsf{Recap} and other methods on Celeb-DF, WildDF, and DFDC datasets when trained on one type of videos of FaceForensics++. }
\vspace{-0.3cm}
\label{crossffone}
\renewcommand{\tabcolsep}{0.6mm} 
\renewcommand{\arraystretch}{1.2}
\begin{tabular}{cccccccccccccc}
\noalign{\hrule height 0.7pt}
     \multirow{2}{*}{Method}
     &\multicolumn{3}{c} {DF}&\multicolumn{3}{c} {F2F}& \multicolumn{3}{c} {FS}&\multicolumn{3}{c} {NT}&\multirow{2}{*}{Avg}\\ 
    \cmidrule(r){2-4} \cmidrule(r){5-7}\cmidrule(r){8-10}\cmidrule(r){11-13}&Celeb-DF &WildDF &DFDC &Celeb-DF &WildDF &DFDC&Celeb-DF &WildDF &DFDC&Celeb-DF &WildDF &DFDC\\
     
              \noalign{\hrule height 0.7pt}


{\textmd{MultiAtt}}&$68.7$&$68.8$&$70.1$&$69.6$& $69.5$&$68.6$&$70.4$&$70.5$&$70.1$&$70.2$&$69.9$&$66.9$&$63.7$ \\
{\textmd{LipForensics}}&$69.3$&$70.2$&$70.8$&$69.1$& $72.4$&$71.4$&$72.3$&$71.9$&$71.8$&$70.9$&$73.2$&$69.8$&$71.1$ \\
{\textmd{Huang et al.}}&$72.9$&$73.9$&$72.8$&$74.2$& $73.3$&$75.8$&$72.7$&$73.0$&$71.9$&$74.8$&$74.1$ &$73.5$&$73.6$\\
\hline
{\textmd{FInfer}}&$68.6$&$69.9$&$67.3$&$68.1$& $69.9$&$66.4$&$70.6$&$66.4$&$67.7$&$70.7$&$69.0$&$69.7$&$68.7$ \\
{\textmd{SSL-AE}}&$73.0$&$72.8$&$\textbf{77.2}$&$78.1$& $72.3$&$78.7$&$80.0$&$77.2$&$74.2$&$75.9$&${75.5}$&$74.1$ &$75.8$\\
{\textmd{OST}}&${76.6}$&$75.2$&${75.7}$&$79.9$& ${78.3}$&$\textbf{79.8}$&$79.2$&$79.3$&$\textbf{80.2}$&$75.3$&$74.8$&$75.2$&$77.5$\\

{\textmd{RECCE}}&$69.7$&$68.3$&$68.0$&$70.5$& $70.0$&$71.1$&$69.7$&$69.3$&$71.1$&$70.1$&$70.4$&$70.2$&$69.9$ \\
{\textmd{MRL}}&$72.9$&$71.3$&$72.2$&$70.6$& $72.5$&$71.2$&$73.1$&$74.0$&$70.5$&$71.4$&$71.2$&$72.4$&$71.9$ \\
 \rowcolor{gray!20}
{\textsf{Recap}}
&$\textbf{77.3}$&$\textbf{75.5}$&$75.4$&$\textbf{80.1}$& $\textbf{79.3}$&$78.9$&$\textbf{81.0}$&$\textbf{79.5}$&$79.8$&$\textbf{76.7}$&$\textbf{75.6}$&$\textbf{76.9}$&$\textbf{78.0}$\\

\noalign{\hrule height 0.7pt}

\end{tabular}
\end{center}
\vspace{-0.6cm}
\end{table*}

Second, to avoid performing experiments on a particular training mode, we change the training mode and conduct other unknown domain detection experiments. Specifically, we implement experiments by selecting one type of FaceForensics++ for training, but testing on other datasets, i.e., Celeb-DF, WildDF, DFDC. Since there is only one type of video for training in experiments, we randomly split the training data into Meta-train and Meta-test with $7:3$. Results in Table \ref{crossffone} illustrate that \textsf{Recap} outperforms previous methods in many scenarios on average. However, it should be noted that SSL-AE\cite{self} and OST \cite{chen2022ost} perform better than ours when testing on DFDC.  Despite these results, it is worth noting that the proposed \textsf{Recap} achieves comparable performance on average.

\subsection{Intra-dataset Detection Performance} To provide a comprehensive assessment of the proposed \textsf{Recap}, we compare \textsf{Recap} with the state-of-the-art methods in the scenario of intra-dataset detection. Specifically, we conduct experiments on $4$ subsets of FaceForensics++ (C23). The training data and testing data of intra-dataset experiments are  from the same subset of FaceForensics++. Table \ref{intra} shows  that most methods perform well in intra-dataset detection.  \textsf{Recap} achieves the highest intra-dataset detection score on Face2Face while having a slight decrease of  $0.7\%$ in average accuracy compared to LipForensics\cite{lips}, which ranks highest on average among the evaluated approaches. 

\begin{table}[!t]
\setlength{\abovecaptionskip}{0.1cm}
\setlength{\belowcaptionskip}{0.1cm}
\tabcolsep=2pt
\begin{center}
\caption{Comparisons of the Intra-dataset evaluation  (AUC (\%)) between \textsf{Recap} and other methods.}
\label{intra}
\renewcommand{\tabcolsep}{3mm} 
\renewcommand{\arraystretch}{1.2}
\begin{tabular}{cccccc}
\noalign{\hrule height 0.7pt}
     {Method}& {DF}&{FS}& {F2F}&{NT}&Avg\\
           
\noalign{\hrule height 0.7pt}

 
 
 
 

 {\textmd{MultiAtt}}&$99.6$&${\textbf{100}}$& ${99.3}$& ${98.3}$&$99.3$ \\
{\textmd{LipForensics}}&${\textbf{99.8}}$&${\textbf{100}}$& $99.3$ & ${\textbf{99.7}}$&${\textbf{99.7}}$\\
{\textmd{Huang et al. }}&$99.6$&${{99.8}}$& $99.5$ & ${98.4}$&$99.3$\\
\hline
{\textmd{FInfer }}&$98.4$&${96.0}$& ${93.5}$ & $94.9$&$95.7$\\
{\textmd{SSL-AE }}&${99.2}$&${99.4}$& $96.0$ & $99.0$&${98.4}$\\
{\textmd{OST}}&${{99.0}}$&${98.8}$& $99.1$ & $95.9$&${98.2}$\\
{\textmd{RECCE }}&$99.7$&${{99.9}}$& $99.2$ & ${98.4}$&$99.3$\\
{\textmd{MRL }}&${{99.2}}$&${{98.1}}$& $97.3$ &$98.6$& ${{98.3}}$\\

  \rowcolor{gray!20}
{\textsf{Recap}}
&${99.5}$&$99.6$&${\textbf{99.6}}$&$97.3$&$99.0$\\

\noalign{\hrule height 0.7pt}
\end{tabular}
\end{center}
\vspace{-0.4cm}
\end{table}

           



\begin{table}[!t]
\setlength{\abovecaptionskip}{0.1cm}
\setlength{\belowcaptionskip}{0.1cm}
\tabcolsep=2pt
\begin{center}
\caption{Ablation study - The detection performance (AUC (\%)) of different masking ratios  on the testing datasets after training on FaceForensics++.}
\label{ratio}
\renewcommand{\tabcolsep}{4mm} 
\renewcommand{\arraystretch}{1.2}
\begin{tabular}{cccc}
\noalign{\hrule height 0.7pt}
     {Mask ratio}& {Celeb-DF}& {WildDF}&{DFDC}\\
           
\noalign{\hrule height 0.7pt}
\textmd{55\%}&$88.2$&${83.0}$ &$79.9$  \\
 
\textmd{65\%}&$89.4$&${82.6}$ &$80.3$  \\
  \rowcolor{gray!20}
\textmd{75\%}&${\textbf{90.1}}$&${\textbf{83.1}}$ &${\textbf{83.2}}$  \\
 
\textmd{85\%}&$89.1$&${82.8}$ &$80.7$  \\
 
 \textmd{95\%}&$88.5$&${82.7}$ &$80.0$  \\
\noalign{\hrule height 0.7pt}

\end{tabular}
\end{center}
\vspace{-0.7cm}
\end{table}

\vspace{-0.1cm}
\subsection{Ablation Study}
\label{ablation}

To implement the ablation study, we conduct experiments by training on FaceForensics++ but testing on the Celeb, WildDF, and DFDC datasets. 

\noindent\textbf{Influence of the masking ratio.} We trained models on the FaceForensics++ dataset with different masking ratios. Note that instead of defining the masking ratio as the ratio of masked area to the entire face, we define the masking ratio as the ratio of masked area to the corresponding ROIs facial parts, as illustrated in Algorithm \ref{algorithm1}.  The reason why we do not use the original definition of mask ratio, that is, the ratio of the mask area to the whole face, is that we focus on ROIs and split faces into three parts and only randomly mask one part of the whole face at one time. We are supposed to focus on the corresponding masked ROIs parts in the masking process rather than the whole face. 

\indent In Table \ref{ratio}, we observe that \textsf{Recap} scales well with the masking ratio of 75\%.  The performance gets a slight drop in the masking ratio of 55\% and 65\% indicating that low masking ratios may hinder learning robust features. When the mask rate is 85\% and 95\%, the detection performance is also degraded. That may be because that high  masking ratio can raise the difficulty of reconstructing faces. If both real faces and fake faces are not reconstructed well, the distinction between them can be reduced. Therefore, we set the masking ratio as 75\%  in the experiments.

\begin{table}
\setlength{\abovecaptionskip}{0.1cm}
\setlength{\belowcaptionskip}{0.1cm}
\tabcolsep=2pt
\begin{center}
\caption{Ablation study - The  detection performance (AUC (\%)) of different mask strategies  on the testing datasets after training on FaceForensics++.}
\label{masks}
\renewcommand{\tabcolsep}{2.2mm} 
\renewcommand{\arraystretch}{1.2}
\begin{tabular}{cccc}
\noalign{\hrule height 0.7pt}
     {Masking strategy}& {Celeb-DF}& {WildDF}&{DFDC}\\
           
\noalign{\hrule height 0.7pt}

\textmd{MAE masking}&$85.1$&$79.9$&$78.3$  \\
 \textmd{VideoMAE masking}&$85.2$&$79.5$&$78.7$  \\
\textmd{Eye}&$89.8$&$82.9$& $80.1$\\

{\textmd{ cheek \& nose}}&$88.9$&$82.1$& $80.5$ \\

{\textmd{Lip}}&$89.5$&$82.2$& $80.9$ \\

 {\textmd{w/o ROIs}}&$89.7$&$82.9$& $82.7$ \\
 
 \rowcolor{gray!20}
 {\textmd{Proposed strategy}}&${\textbf{90.1}}$&${\textbf{83.1}}$ &${\textbf{83.2}}$ \\
\noalign{\hrule height 0.7pt}
 
\end{tabular}
\end{center}
\vspace{-0.6cm}
\end{table}

\noindent\textbf{Influence of the masking strategy. }We modify the masking strategies of MAE \cite{imgmae} to improve the generalization. To evaluate the effectiveness of the improved  masking strategy, we compare the proposed masking strategy with masking strategies of MAE and VideoMAE. Furthermore, since the modified strategy randomly selects parts to mask, evaluating the effects of  different masked parts is important. To analyze the effectiveness of the ROIs, we compare the proposed strategy with the masking strategy that does not focus on ROIs. We trained models on
the FaceForensics++ dataset with different masking strategies.

\indent The results of $1^{st}$, $2^{nd}$, and  $7^{th}$ lines  in  Table \ref{masks} demonstrate that modifying the  masking strategies of MAE \cite{imgmae} and VideoMAE \cite{videomae} can improve the detection performance. The results in the $3^{rd}$, $4^{th}$ and $5^{th}$  lines, which represent methods that mask eye areas, cheek and nose areas, and lip areas, respectively, show a performance degradation compared to the proposed strategy.  That is, random masking a part of all facial parts is more conducive to extracting robust features than that masking a certain part only.  Moreover, the results of the $6^{th}$ line and $7^{th}$ lines show that the proposed masking strategy that focuses on ROIs achieves better performance than the  masking strategy without ROIs. The reason  is that   the model can better capture the differences between real and fake videos by masking patches in these ROIs, as fake videos typically lack 
consistency. Therefore, the proposed masking strategy illustrated in Algorithm \ref{algorithm1} is effective in detecting Deepfake videos.

\begin{table}
\setlength{\abovecaptionskip}{0.1cm}
\setlength{\belowcaptionskip}{0.1cm}
\tabcolsep=2pt

\begin{center}
\caption{Ablation study - Effects of MAE, VideoMAE, Recovering stage, Mapping stage,  meta-learning.}
\label{twostream}
\renewcommand{\tabcolsep}{1.5mm} 
\renewcommand{\arraystretch}{1.2}
\begin{tabular}{cccc}
\noalign{\hrule height 0.7pt}
     {}& {Celeb-DF}& {WildDF}&{DFDC}\\
\noalign{\hrule height 0.7pt}
{\textmd{MAE}}&$75.3$& $72.1$&$70.2$ \\
 {\textmd{VideoMAE}}&$76.2$& $73.3$&$71.0$ \\


    \textmd{w/o Recovering stage}&$87.8$&$80.5$& $80.1$  \\

\textmd{w/o Mapping stage}&$84.6$& $79.3$&$79.2 $\\
 
{\textmd{w/o Meta-learning}}&$88.4$& $82.0$&$80.6$ \\
  {\textmd{MAE + Mapping stage}}&$81.6$& $76.7$&$74.3$ \\
   {\textmd{VideoMAE + Mapping stage}}&$82.5$& $77.2$&$75.0$ \\
     {\textmd{RECCE + Mapping stage}}&$80.6$& $76.2$&$75.4$ \\ 
 \rowcolor{gray!20}
{\textmd{Recap}}&${\textbf{90.1}}$&${\textbf{83.1}}$ &${\textbf{83.2}}$ \\
\noalign{\hrule height 0.7pt}
\end{tabular}
\end{center}
\vspace{-0.6cm}
\end{table}

\noindent\textbf{Influence of MAE and VideoMAE.} 
We compare the detection performance of the  \textsf{Recap} with the original MAE and VideoMAE methods for Deepfake detection. The results are shown in the $1^{st}$ and $2^{nd}$ line of Table \ref{twostream}. The detection performance of the original MAE and VideoMAE is lower than that of \textsf{Recap}, demonstrating the effectiveness of the modifications in \textsf{Recap}. 

\noindent \textbf{Influence of Recovering stage and Mapping stage.}   To validate the performance of each stage, we compare  the performance of a single stage with that of both stages combined. The results are shown in the $3^{rd}$ and $4^{th}$ lines of Table \ref{twostream}. We can see that removing either the Recovering stage or the Mapping stage degraded the detection performance, as each stage plays a crucial role in Deepfake detection. Combining both stages improve the performance by magnifying the distinction between real and fake videos.

\noindent \textbf{Influence of meta-learning. }We remove the meta-learning module to carry out experiments, and the results are shown in the $5^{th}$ line of Table \ref{twostream}. The results in the $5^{th}$ and $9^{th}$ lines show that the method without meta-learning achieves worse results than the proposed  \textsf{Recap} with meta-learning. The meta-learning approach simulates cross-domain detection in the training phase, improving  detection performance. 



\begin{figure}[!t]
 \centering
\includegraphics[width=1\linewidth]{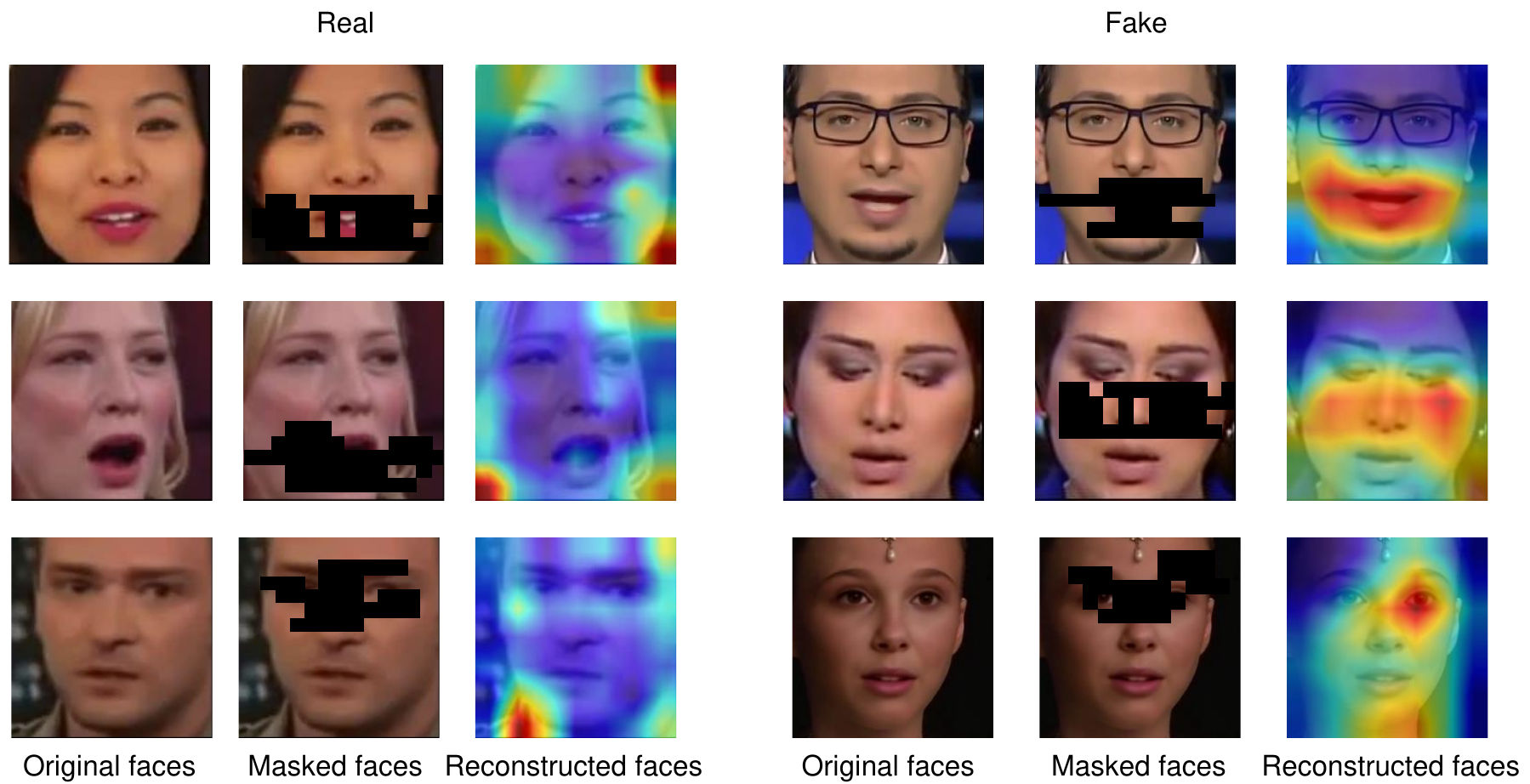}
   \vspace{-0.5cm}
\caption{The activations of the masked and recovered faces  indicate that ROIs recover more easily for real faces than for fake ones.}
 \label{fig3mask}
\vspace{-0.4cm}
\end{figure}


\begin{figure}[!t]

\centering
\subfigure[Before mapping]
{
\includegraphics[width=3.9cm]{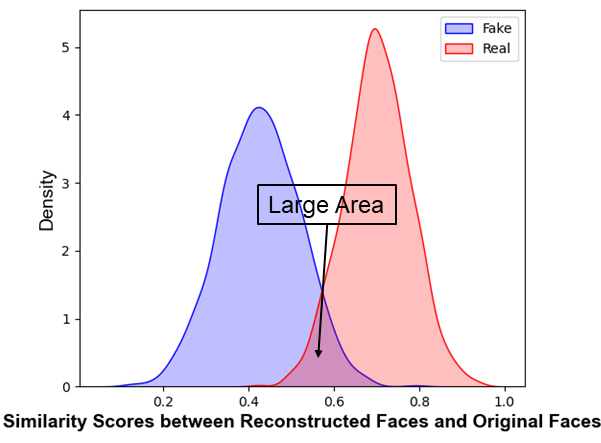}
\label{fig4a}
}
\subfigure[After mapping]
{
\includegraphics[width=3.9cm]{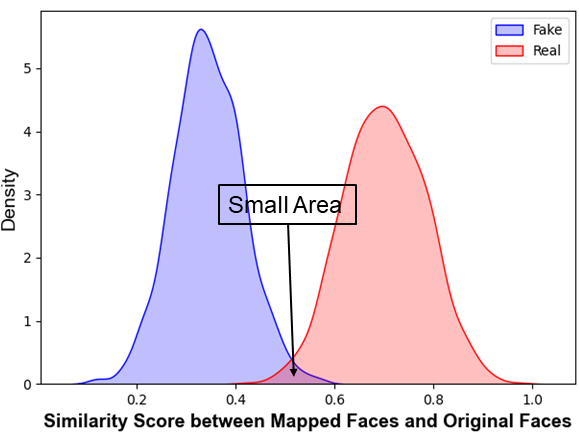}
\label{fig4b}
}
\vspace{-0.4cm}
\caption{The cross area after mapping becomes smaller, indicating that the difference between real and fake videos is enlarged after mapping.}
\label{figcor}
\vspace{-.2in}
\end{figure}

\noindent \textbf{Visualize the activation. } To gain insight into the features extracted by our proposed  \textsf{Recap}, 
we use the CAM function wrapped in \texttt{pytorch\_grad\_cam} to show the activations. This involves masking and reconstructing faces, and then visualizing the activations of recovered faces and mapped faces. For the recovered faces, as shown in Fig. \ref{fig3mask}, the activation areas of real and fake faces are different. the masked areas of the reconstructed fake faces are activated, exposing the inconsistencies of fake faces. In contrast, the consistency of real faces guides the reconstruction, making it difficult for the network to extract abnormal inconsistencies. Therefore, activation areas are present in the background areas rather than the masked areas.

\subsection{Gain of the Mapping Stage}
MAE and VideoMAE  reconstruct images and videos for pretraining ViT, and RECCE reconstructs faces for Deepfake detection. We apply the Mapping stage in MAE, VideoMAE, and RECCE and show the results in the $6^{th}$, $7^{th}$ and $8^{th}$ lines of Table \ref{twostream}. Compared  to the results without leveraging reconstructed results for mapping, it shows that combining the Mapping stage with MAE, VideoMAE, and RECCE can improve the detection performance. Moreover, we present the distributions before and after the mapping stage. Specifically, we utilize Euclidean distance to calculate pixel similarity scores between reconstructed faces and original faces. The similarity scores of normalization are shown in  Fig. \ref{fig4a}. We also calculate pixel similarity scores between mapped faces and original faces, and show the similarity scores of normalization  in  Fig. \ref{fig4b}. It is obvious that real faces and fake faces have different distributions. Furthermore, the cross area of  Fig. \ref{fig4b} is larger than that of  Fig. \ref{fig4a}, which shows that the real and fake videos after mapping are easier to distinguish, thus improving the detection performance.

\section{Conclusion}
This paper focuses on the detection of Deepfakes, particularly in identifying Deepfake videos with  unpredictable tampered traces. By focusing equally on all facial parts rather than relying on specific facial parts, our two-stage model can learn unspecific facial consistencies and robust representations that only exist on real faces.  In the Recovering stage, the model is trained to recover faces from partially masked ROIs on the face, which helps the model learn the facial part consistencies of real videos. In the Mapping stage, the recovered real faces are enforced to be mapped successfully while recovered fake faces are mapped failingly so that to maximize the differences between real and fake videos.  Extensive experiments illustrate the robustness and generalizability of \textsf{Recap} on benchmark datasets.


\bibliography{aaai24}

\end{document}